\documentclass[11pt]{article}
\usepackage{geometry}
\usepackage{coling2020}
\usepackage{times}
\usepackage{url}
\usepackage{latexsym}
\usepackage{microtype}
\usepackage{graphicx}
\usepackage{xcolor}
\usepackage{amsmath,bm}
\usepackage{makecell}
\usepackage{floatrow}
\usepackage{appendix}
\usepackage{enumitem}
\usepackage{hyperref}
\usepackage{titlesec}
\usepackage{lipsum}
\usepackage{siunitx}
\usepackage{caption}
\usepackage{floatrow}

\titlespacing{\paragraph}{%
  0pt}{
  0pt}{
  1em}%
  
\newfloatcommand{capbtabbox}{table}[][\FBwidth]

\colingfinalcopy 

\title{ANA\thanks{~~\url{https://www.amii.ca/ana-automated-nursing-agent/}} ~ at SemEval-2020 Task 4:\\ mUlti-task learNIng for cOmmonsense reasoNing (UNION)}

\author{Anandh Perumal, Chenyang Huang, Amine Trabelsi, Osmar R. Za\"{\i}ane \\
 Alberta Machine Intelligence Institute, University of Alberta \\
  {\tt \{anandhpe, chenyangh, atrabels,zaiane\}@ualberta.ca} }

\begin{document}
\maketitle
\begin{abstract}
In this paper, we describe our mUlti-task learNIng for cOmmonsense reasoNing (UNION) system submitted for Task C of the SemEval2020 Task 4, which 
is to generate a reason explaining why a given false statement 
is non-sensical. 
However, we found in the early experiments that simple adaptations such as fine-tuning GPT2 often yield dull and non-informative generations (e.g. simple negations). 
In order to generate more meaningful explanations, we propose UNION, a unified end-to-end framework, to utilize several existing commonsense datasets so that it allows a model to learn more dynamics under the scope of commonsense reasoning. 
In order to perform model selection efficiently, accurately and promptly, 
we also propose a couple of auxiliary automatic evaluation metrics 
so that we can extensively compare the models from different perspectives. Our submitted system not only results in a good performance in the proposed metrics but also outperforms its competitors with the highest achieved score of 2.10 for human evaluation while remaining a BLEU score of 15.7. 
Our code is made publicly available at GitHub \footnote{\url{https://github.com/anandhperumal/ANA-at-SemEval-2020-Task-4-UNION}}.

\end{abstract}

\section{Introduction}
Common sense reasoning is one of the long-standing problems in natural language understanding. 
Previous work on modeling common sense knowledge deals mainly with indirect 
tasks
such as co-reference resolution \cite{winogrande}, 
or selecting the plausible situation based on the given subject or scenario \cite{SWAG}.

In this paper, we present our system that we devised to tackle Task C, Explanation (Generation), of the SemEval 2020 Task 4 - Commonsense Validation and Explanation (ComVE).
Given a false or non-sensical statement, the task consists of generating the reason why the given statement does not make sense. We propose a mUlti-task learNIng for cOmmonsense reasoNing (UNION).
It combines datasets including ComVE \cite{ComVE}, OpenBook \cite{OpenBook}, Common sense Explanation (CoS-E) \cite{COSE} and Open Mind Common Sense (OMCS) \cite{omcs} in a multi-task framework. 
%
The backbone of UNION is a large pre-trained language model -- GPT2.


We compare the proposed system to different baselines
and report a significant improvement in BLEU score and other evaluation metrics. 
%
Our proposed model achieves a human evaluation score of 2.10, which ranked first on the final leader board for Task C of SemEval 2020 Task 4.
In our initial submission, we used ComVE, CoS-E, and OpenBook datasets for training. In that case, a BLEU score of 15.7 is achieved. Pretraining the model with OMCS dataset further improved the BLEU score by 0.7. In addition, we show some of our generations in the appendix.

\section{Background and Related Work}

The
common sense 
reasoning in machines
has been one of the most challenging problems, and a major critical missing component of Artificial Intelligence (AI). It could be helpful in various aspects of day-to-day life \cite{11}. For any AI system to generate commonsense reasoning, it needs to understand and build a representation of the given situation \cite{muellercommonsense}. 
For instance, let us consider the following statement 
``You will never find a dog that likes to eat meat." 
Humans
may have intuitively built a knowledge graph 
containing the facts
that ``dogs are carnivorous" and ``a carnivorous is a meat-eater". 
Thus, based on it, they may conclude that the first statement
is a false statement.
Similarly, for a machine to answer questions about general situations or facts, a knowledge graph or formal logic system could be beneficial. Early work by John McCarthy \shortcite{mccarthy} proposed a system that uses formal logic for commonsense reasoning. Thus, the system can induce commonsense reasoning given all the possible axioms 
about
the studied
domain or 
the world. 
However, 
it
is not feasible to generate all the reasonable axioms about the world. 
This
gave rise to several other logic-based approaches for commonsense reasoning and numerous work with the aim of creating huge logic-based ontologies, e.g., situation calculus \cite{situationcalculus}, YAGO \cite{yago}, DBpedia \cite{dbpedia}, Event2mind \cite{Event2mind}. 
However, commonsense reasoning over a particular knowledge often acts as a lookup table as it lacks sufficient semantics to formulate a complete sentence as a reason.
Other models
\cite{COSE,BERT,qads1,taylor1953cloze} 
have
achieved competitive performance in question answering tasks given a comprehension or a document as a source of information. All these models help to 
simulate
common sense reasoning in the language model. However, they still rely on a passage as a source of information, which makes 
them
limited to specific domains.

An AI system 
which
processes common sense like humans, 
in the context of question answering or dialogue generation,
should not depend on a source of information each time it produces a response to a query. Instead, it should learn and generate a response from the previously learned information.
Thus, our proposed model does not rely on any other source of information during inference other than its prior learned knowledge. We train UNION on multiple commonsense datasets ComVE, CoS-E, and OMCS.

\section{Proposed systems}
\label{proposedSystem}
The language generation task is to model the conditional probability $P(Y|X)$, where $X$ and $Y$ are both sequences of words. In this work, we use the 36-layer decoder-only transformers based on the architecture of the Generative Pre-trained Transformer (GPT2) \cite{GPT2} as the backbone of the language generation task for all the presented models.
%
%
The decoder only transformer architecture is similar to the original decoder transformer architecture proposed by Vaswani et al. \shortcite{vaswani2017attention}. The main difference 
is that in the decoder-only transformer case, the decoder component does not have a multi-head attention for the encoder input. All our models are initialized with the large weights of the pre-trained language model GPT2.
In this section, we
review
the baseline models
and the architecture of our proposed UNION model.
\subsection{Baseline models}
\paragraph{Language generation baseline:} \label{baseline}


For Task C Explanation (Generation), the model needs to generate the reason why a given false statement is against common sense. 
The data is given in the format of $\{X^{i}, \bm{Y}^{i}\}$, where each statement $X^{i}$ is paired with a three possible explanations $\bm{Y}^{i} = (Y_{1}^{i}, Y_{2}^{i}, Y_{3}^{i})$.

%
%
We formulate Task C as a sequence-to-sequence generation task. A false statement $X^{i}$  is considered as  the source to the language model while one of the explanations from $\textbf{Y}^{i}$  becomes the target.
Hence, as a first step, we transform our dataset format. 
It is no longer of the format $(X^{i}, Y_{1}^{i}, Y_{2}^{i}, Y_{3}^{i})$, but of the format $(X^{i}, Y_{j}^{i})$, where  $Y_{j}^{i} \in \textbf{Y}^{i}$  for $j=1,2,3$. 
As a second step , we train our baseline language model GPT2 with the new dataset format. 
The final goal is to
estimate the conditional probability distribution $P(Y_{j}^{i} | X^{i})$ of the target statement  $Y_{j}^{i}$ given the source statement  $X^{i}$, where $j=1,2,3$.

\paragraph{Multi-task learning baseline:}
\label{MTLBaseline}

We train a language model in the Multi-task learning (MTL) setting. The model has two heads, one for language model and another one for the classification task.
Caruana \shortcite{caruana1997multitask} has briefly expressed the advantage of MTL: ``\textit{MTL improves generalization by leveraging the domain-specific information contained in the training signals of related tasks}".
Moreover, Raffel et al. ~\shortcite{T5}, Liu et al. ~\shortcite{MTDNN}, and Devlin et al.~\shortcite{BERT} have shown that training language models using an MTL technique helps the language model to learn and generalize better.

We chose to train
an MTL model 
using ComVE's Task B and Task C datasets because of their similarity.
Task B of ComVE is a multi-choice classification problem. For each false statement $X^{i}$,  
there exist three plausible explanations from $\mathbf{\hat{Y}^{i}} = (\hat{Y}^{i}_{1}, \hat{Y}^{i}_{2}, \hat{Y}^{i}_{3})$.
Although, only one of these 
is correct,
they all have the
same syntactic structure and similar wording, and  only 
differ
by few words.
Moreover, all of the correct explanations 
present in Task B
are
a subset of explanations $\mathbf{Y}$ provided in Task C (Generation). 
Thus, we hypothesized that adding this task in an MTL setting would help the model  
better discriminate between a valid and an invalid explanation, and consequently 
learn to recognize the subtle differences allowing it to generate better valid reasons. Moreover, training on more similar examples to those in Task C would
make the generated text
more aligned with the syntactic structure and keywords of the task's data. 

We convert
the dataset 
format
from a multi-label 
to a binary classification 
format $(X^{i}, \hat{Y}^{i}_{k}, b)$,
where $b$ is a binary label (correct or not), and $k=1,2,3$. 
The Task B and C 
have
the same set of false statements $X$ as input. Thus, we pair the dataset according to the false statement. The new data format is the following:  $\{(X^{i}, Y^{i}_{j}) , (X^{i}, \hat{Y}^{i}_{k}, b)\}$. We train the language modeling and the classification heads in parallel. The loss is computed by taking the summation of both heads' losses.

\subsection{Why UNION?}
\label{sec:unite}
A language model is a probability distribution learned over the sequence of tokens in the training corpus. Thus, the performance of the model strongly depends on the quality of the data used for its training.
When we investigated the explanations generated by the baseline and MTL models, it seemed that these tended to often negate the given input statement.
Although this
may still make them valid explanations, 
it
did not necessarily make them
express
reasoned explanations on why the input is a false statement. 
Part A
of 
Table
\ref{tab:informative}
provides examples of two valid  explanations 
where
the first is just a negation of the input whereas the second 
is a better systematic explanation of why the given statement is false.
The root cause for generating a simple 
negation of a statement over
a reasoned 
explanation is the ComVE dataset itself.
More than one-third of 
it contains
negating statements, which most probably signals the model to learn to negate any given input.
To deal with 
this
issue, 
we may either remove the explanations that are negation to the false statement or increase the dataset by adding better explanations to it.
Since the dataset for explanation generation is limited, deleting any explanations may have a negative impact, while creating a new dataset is a tedious task in itself. Therefore, we resorted to using other commonsense related datasets, CoS-E, OpenBook,  and OMCS ,  along with the ComVE dataset, to train the language model to generate better responses. 
These additional datasets treat different issues.
OpenBook is a question answering dataset related to science facts. CoS-E is for general commonsense question answering. OMCS contains common knowledge facts. 
Training all four together leads to two main difficulties or questions to answer:
\begin{enumerate}[topsep=0pt,itemsep=-1ex,partopsep=1ex,parsep=1ex]
\item Each dataset is 
related to
 a different issue while 
 our main task is to generate an explanation for a false statement. How can we force the model to generate an explanation response that is specific to the false statement and not any random generic statement related to it?
\item Each dataset has a different number of classification choices. How do we train all of them together in an MTL setting?
\end{enumerate}
Our proposed architecture, described in Section \ref{union}, solves the first problem with the help of a contextual keyword and the second problem by merging all the classes.

\begin{figure}
\RawFloats
\begin{minipage}{0.45\linewidth}
    \includegraphics[width=.98\textwidth]{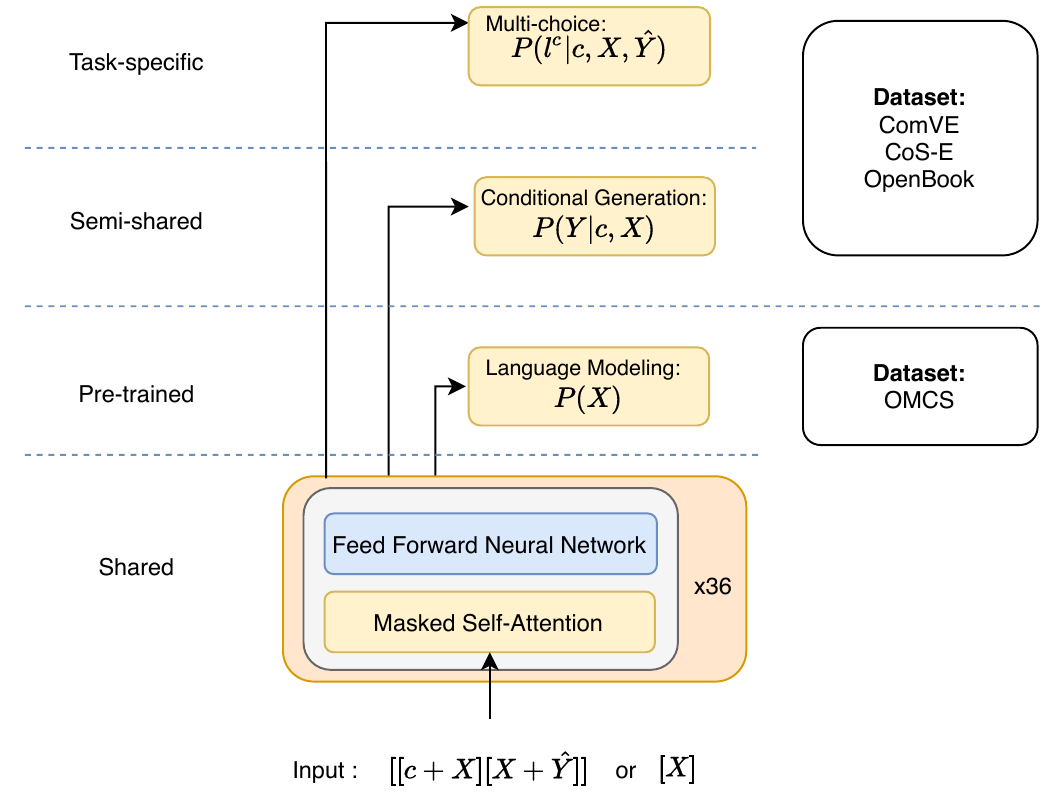}
    \caption{UNION Architecture}
\label{fig:unionarch}
\end{minipage}
\hfill
\begin{minipage}{0.45\linewidth}
\resizebox{.98\textwidth}{!}{
  \begin{tabular}{l|l}
        \hline \hline
        \makecell{\textbf{False} \\ \textbf{Statement}} &  We use book to know the time \\
        \hline
        \makecell{\textbf{Referential} \\ \textbf{Reasons}} & a) A book is used to study \\
        & b) A book does not have the \\ & ability to show what time it is. \\
        & c) Books don't tell the time \\
        \hline
        \makecell{\textbf{Generated} \\ \textbf{Explanation}} & Book is not a timekeeping device.    \\      
        \hline \hline
    \end{tabular}}
    \captionof{table}{UNION model generated explanation}
    \label{tab:bleuexample}
\end{minipage}
\end{figure}


\subsection{UNION} \label{union}



The backbone of the mUlti-task learNIng for cOmmonsense reasoNing (UNION) architecture in Figure~\ref{fig:unionarch} is a decoder-only transformer, as described above. 
The UNION model is categorized into four major layers, shared, pre-trained, semi-shared, and task-specific layer.  The first layer is a shared layer among all the other following layers. The second layer is a language modeling head. It is used to pre-train the UNION model with the OMCS dataset before training with ComVE, CoS-E, and OpenBook. The OMCS dataset has over a million statements as facts and common knowledge, e.g., ``You are likely to find a shelf in a cupboard.". The pre-training helps the model to learn about the general facts and knowledge about the world.
The third layer is semi-shared language modeling (SSLM) head shared among data from ComVE, CoS-E, OpenBook. It is used to train the explanation generation for all three datasets. 

We use a  contextual keyword ($c$)
to solve the first issue 
described in Section
\ref{sec:unite}, i.e., to generate a response or an explanation that is specific to a particular dataset.
The contextual keyword is a unique token for each dataset. During training, we condition the $Y$ on $(c, X)$. Therefore, we no longer estimate the  $P(Y|X)$, but we estimate the $P(Y|c, X)$, while 
during
inference, we condition the false statement of the ComVE dataset to the particular contextual keyword to generate an explanation.
We call this layer semi-shared because we share the same language modeling head layer for ComVE, CoS-E, and OpenBook 
while we also condition
the generation for each given task
based on the contextual keyword. 

Finally, we have a task-specific layer. The task-specific layer is used for multi-class classification (MC), and the same contextual keyword for the SSLM head layer is used in the MC layer too. The final classification layer size is twelve (first three classes for ComVE, four for OpenBook, and five for CoS-E). Combining all the classes helps to mitigate the second problem described in the previous section of the 
difference in
the number of classes
between
datasets.

Therefore, during training and inference, with the help of the contextual keyword,  we ignore the labels which are not relevant for the particular dataset by assigning zero and normalizing the probability distribution over the remaining relevant labels. For example, during the training of the ComVE dataset using the contextual keyword, we are assigning zero probability to the labels from four to twelve (i.e., those related to OpenBook and CoS-E), and the probability distribution is normalized over the first three labels. 
So, we estimate the $P(l^{c} | c, X, \hat{Y})$, where $\hat{Y}$ is a concatenation of all the 
potential
explanations for $X$, $l^{c}$ is the label, and $c$ is the contextual keyword.


\section{Experiments}
\subsection{Evaluation metrics}
The BLEU score \cite{papineni2002bleu} is widely used in tasks such as machine translation \cite{MTBleu1,MTBleu2}. It calculates the overlapping between the candidates and the reference text. However, similar to dialogue generation, 
our experimental results show that the BLEU score is not an ideal measurement since it does not tolerate diversity.
For instance, as shown in Table~\ref{tab:bleuexample}, 
the generated explanation example
``Book is not a timekeeping device.'' has 
little overlap with the reference but it should still be considered as a good generation. 
As a remedy, the submitted systems are 
evaluated by human evaluations.
The BLEU score and the human evaluation are carried out by the task organizers, where the human evaluation score is based on the agreement between three reviewers. 
In order to better evaluate 
the performance of 
the proposed systems and conduct ablation studies, we
use several auxiliary
metrics to assess the quality of the  explanations  generated by various models.
\begin{table}[ht]
\centering
\resizebox{.8\textwidth}{!}
    {
        \begin{tabular}{l|c|c|c|c|c|c} \hline \hline
            \textbf{Models}	& \textbf{BLEU} & \textbf{PPL - Gen.} & \textbf{PPL - Trg.} & \textbf{EA} & \textbf{UNI} & \textbf{Length} \\ \hline
            \textbf{Baseline}		 & 10.36	& 970.05	& 495.35	& 0.86I	& 3.55 $\pm$ 1.77	& 5.5  $\pm$ 1.97	   \\
            \textbf{Baseline + MTL}	 & 12.4	 & 357.59	& 331.22	& 0.93	& 3.31 $\pm$ 1.68	& 5.59 $\pm$ 1.89	   \\
            \textbf{UNION w/o CoSE}	& 13.28	& 62.89	 & 238.64	& 0.96	& 5.82 $\pm$ 2.10	& 8.51 $\pm$ 2.08	   \\
            \textbf{UNION w/o OpenBook} & 13.75	& 142.19	& 260.38	& 0.95	& 4.29 $\pm$ 1.87	& 6.46 $\pm$ 2.19	   \\
            \textbf{UNION w/o OMCS}	& 15.7	 & 194.66	& 243.83	& 0.94	& 4.29 $\pm$ 1.79	& 6.41 $\pm$ 2.06	   \\
            \textbf{UNION}	 & 16.36	& 135.1	 & 212.1	 & 0.97	& 4.53 $\pm$ 2.05	& 6.59 $\pm$ 2.3	\\	   \hline \hline
        \end{tabular}
    } 
    \caption{Ablation study results on different proposed models}
    \label{tab:result}
\end{table}

\paragraph{Perplexity:}
We first use perplexity to measure the [grammatical/syntactical] correctness of the generated text. Particularly, we suggest two variants of perplexity: the general perplexity (\emph{ppl-gen}) and the target corpus perplexity (\emph{ppl-trg}). 
We measure \emph{ppl-gen} by using the GPT2 language model (LM) head directly, as it is pre-trained on large scale corpus (Reddit, Wiki). It gives an indication on the fluency of 
the generated content 
in general.
In the meantime,  we also want to assess how well the generations can be in terms of fitting into the dialect of the target corpus. We achieve this by  training an n-gram language model with Kneser-Ney smoothing\footnote{\url{https://github.com/kpu/kenlm}}.
The n-gram language model is trained based on Task B and C datasets.
\paragraph{Informative:}
As shown in Table~\ref{tab:informative},  the generations produced by the baseline models are often derived from the false statement $X^i$
by changing very few words.  To measure the impact of the change in the generated responses, we propose two auxiliary metrics: 
\emph{Estimated Approval (EA)} and \emph{Uniqueness (UNI)}.

\begin{table}[ht]
    \centering
    \resizebox{.8\textwidth}{!}{
    \begin{tabular}{l|l}
    \hline \hline
         False Statement &       Explanation\\
    \hline
        A) The chocolate cried. & 1) Chocolate \textcolor{green}{doesn't} cry. \\
        & 2) Chocolate is an inanimate, non human thing and cannot cry.\\
        
        B) Sugar is used to make coffee \textcolor{red}{sour}. & 1) Sugar is used to make coffee \textcolor{green}{sweet}. (UNION)\\
        & 2) Sugar is \textcolor{green}{not} used to make coffee sour. (Baseline)\\
\hline \hline
    \end{tabular}}
    \caption{Examples from ComVE explanation (Generation), and UNION and Baseline models}
    \label{tab:informative}
\end{table}
The UNI and Length metric measures 
the added amount of information of an
answer.
UNI calculates the number of tokens that are not present in the given input and Length measure the length of each explanation. 
A high UNI and Length value suggests more diverse keywords used and potentially more informativeness.
On the other hand, an entirely irrelevant text may also achieve a high UNI and Length score. Therefore, in addition to UNI, we train Estimated Approval (EA), an external discriminator, 
whether the generations are valid or not.  
EA is a binary classification model that is initialized with pre-trained BERT weights \cite{BERT}.
We use sub-tasks B and C datasets to fine-tune our EA classifier to estimate the $P(b| X^{i}, Y^{'i})$, where $Y^{'i} \in  \cup({\mathbf{Y}, \mathbf{\hat{Y}}})$. The EA score indicates the average number of explanations generated by the language model that are valid explanations according to the EA.


\subsection{Analysis}


Table \ref{tab:result}  summarizes the results obtained by the various models that we have tried for this task. 
The difference between \emph{UNION w/o CoS-E}, \emph{UNION w/o OpenBook}, \emph{UNION w/o OCMS}, and  \emph{UNION} is only the training datasets while the architecture for all the model remains the same. The perplexity, EA score, and the average length of the generations by \emph{UNION w/o CoS-E} are better than \emph{UNION w/o OpenBook}. 

It is important to note that
the CoS-E dataset is an open-ended common sense question answering.
The ComVE dataset, however, has been constructed by annotators who were influenced by ConceptNet
to generate
false statements. 
The ConceptNet is a semantic graph of commonsense knowledge developed from the OMCS dataset. 
Thus, training the UNION model with the OMCS dataset 
leads to
better results than the \emph{w/o OCMS model}. 
The OpenBook dataset is related to scientific facts, which is similar to the OMCS dataset. Thus, the \emph{UNION w/e CoS-E} performs better than \emph{UNION w/o OpenBook} while the UNION model performs better than all the other models.
The initial model we submitted to the SemEval2020 Task 4
is \emph{UNION w/o OCMS}.  
It achieved a BLEU score of 15.7 and a human evaluation score of 2.10 (Ranked 1st). Later, when training the model with the OMCS dataset, the BLEU score of the model increased by 0.7. 
Examples of generated explanations from all models are provided in appendix \ref{app1}.

\section{Conclusion}
In this research, we propose a UNION model for multi-tasking on a couple of commonsense related datasets, which helped our UNION model to achieve state-of-the-art in the ComVE by achieving the highest human evaluation score. 
In addition, we propose several auxiliary metrics to better evaluate different models for commonsense response generation tasks without human evaluations. 
In the future, we would like to explore other possible areas of commonsense reasoning like quantitative reasoning, logic puzzles, and visual common sense reasoning.

\bibliographystyle{coling}
\bibliography{semeval2020}

\appendix
\section{Examples of generations}
\label{app1}

\noindent
\textbf{False statement :} The Pokemon caught Ash \\
\textbf{Baseline :} a pokemon is not a bird	\\
\textbf{Baseline + MTL :} pokemon are not animals \\	
\textbf{UNION w/o Cose :} a pokeball is used to catch a pokemons	\\
\textbf{UNION w/o OpenBook : }pokemon are not alive	\\
\textbf{UNION w/o OCMS :} pokemon are not alive	\\
\textbf{UNION :} Ash is a human and cannot be caught by a pokemon. \\

\noindent
\textbf{False statement :} You should iron your clothes before you wash them \\
\textbf{Baseline :} ironing clothes is not the same as washing	\\
\textbf{Baseline + MTL :} ironing clothes is not a good thing to do\\	
\textbf{UNION w/o Cose :} ironing clothes before washing them will make them dirty. \\	
\textbf{UNION w/o OpenBook :} ironing clothes before washing them will make them dirty. \\
\textbf{UNION w/o OCMS :} ironing your clothes before washing them will make them cleaner	 \\
\textbf{UNION :} you should wash your clothes before you iron them\\

\noindent
\textbf{False statement :} I was on date with lion yesterday.\\
\textbf{Baseline :} lion was not on date with girl	\\
\textbf{Baseline + MTL :} lion is not a date.	\\
\textbf{UNION w/o Cose :} lion is not a human to be on date with.	\\
\textbf{UNION w/o OpenBook :} lion is a wild animal and cannot be on a date. \\	
\textbf{UNION w/o OCMS :} lion is not a human to date.	\\
\textbf{UNION :} Lions are wild animals and are not meant to be on dates. \\

\noindent
\textbf{False statement :} He broke his leg and went to get a burger \\
\textbf{Baseline :} burgers are not dangerous	\\
\textbf{Baseline + MTL :} burgers are not broken.\\	
\textbf{UNION w/o Cose :} you don't get a burger from breaking your leg.	\\
\textbf{UNION w/o OpenBook :} a burger is not a medical treatment	\\
\textbf{UNION w/o OCMS :} burgers are not used to heal broken legs.\\	
\textbf{UNION :} Break your leg and you will get a cast.\\

\noindent
\textbf{False statement :} People use cars to fly.\\
\textbf{Baseline : } cars are not the most efficient \\	
\textbf{Baseline + MTL :} cars are not used for flying	\\
\textbf{UNION w/o Cose :} cars don't have the ability to fly.	\\
\textbf{UNION w/o OpenBook :} cars don't fly.	\\
\textbf{UNION w/o OCMS :} cars cannot fly.	\\
\textbf{UNION :} Cars are not designed to fly.\\

\end{document}